\documentclass[letterpaper,journal,twoside]{IEEEtran}
\usepackage{cite}
\usepackage{array}
\usepackage{makecell}
\usepackage{graphicx} \graphicspath{{./figures/}}
\usepackage{amsmath,amsfonts}
\usepackage{url}
\usepackage{doi}
\usepackage{import}
\usepackage{multirow}
\usepackage{balance}
\usepackage{xcolor}
\usepackage{booktabs}

\usepackage{tikz}
\usetikzlibrary{calc}
\usepackage{xfp}
\usepackage{xparse}
\NewDocumentCommand\slashed{O{0} O{0} O{1.0} O{30} O{black} m}{%
  \begin{tikzpicture}[baseline=(X.base)]
    \node[inner sep=0pt, outer sep=0pt] (X) {\(#6\)};
    \draw[#5, thick]
    let
      \p1 = (X.north east),
      \p2 = (X.south west),
      \n1 = {\x1 - \x2}
    in
        ($ (X.center) + (#1, #2) + (\fpeval{-0.5*#3*\n1}pt, \fpeval{-0.5*#3*\n1*tan(#4*pi/180)}pt) $)
          --
        ($ (X.center) + (#1, #2) + (\fpeval{0.5*#3*\n1}pt, \fpeval{0.5*#3*\n1*tan(#4*pi/180)}pt) $);
    \end{tikzpicture}
}

\begin{document}

\title{Design and Characterization of a Limb-Encircling Actuator}

\author{
    Japmanjeet~Singh~Gill, Gray~C.~Thomas, Nikko~Van~Crey, Leo~Medrano, and Elliott~J.~Rouse

    \thanks{This work was supported by the D. Dan and Betty Kahn Foundation. The work of Nikko Van Crey was supported by the National Science Foundation Graduate Research Fellowship (Grant No. DGE-1841052). \textit{(Corresponding author: Elliott J. Rouse)}}

    \thanks{Japmanjeet Singh Gill and Nikko Van Crey are with the Department of Robotics, University of Michigan, Ann~Arbor, MI~48109, USA (e-mail: jjsgill@umich.edu)}

    \thanks{Gray C. Thomas is with the Department of Robotics, and also with the Department of Electrical Engineering and Computer Science, University of Michigan, Ann~Arbor, MI~48109, USA.}

    \thanks{Leo Medrano and Elliott J. Rouse are with the Department of Robotics, and also with the Department of Mechanical Engineering, University of Michigan, Ann~Arbor, MI~48109, USA (e-mail: ejrouse@umich.edu).}
}

\maketitle

\begin{abstract}
    Lower-limb powered exoskeletons have demonstrated substantial improvements in mobility and function, but most designs place actuation components lateral to the legs or remotely at the waist or back. These configurations often extend beyond the body’s natural envelope, making devices intrusive in everyday use and potentially limiting societal adoption. We posit that rethinking actuation geometry could enable exoskeletons that conform more closely to the body. Here, we explored an actuation layout in which the actuator encircles the limb in a plane orthogonal to the limb axis, potentially reducing its spatial footprint around the body. We developed the Limb-Encircling Actuator (LEA) and characterized its electromechanical properties using a custom-built testbed. The LEA also features a novel radial bearing layout with potential as a lightweight or lower-cost alternative to traditional large-diameter bearings. The actuator achieved a continuous torque density of 7.5~Nm/kg with a mass of 894~g. Despite this high torque density and innovative layout, the system remained difficult to contain close to the body. These results highlight opportunities and challenges in limb-encircling actuation and provide insights into torque-dense exoskeleton designs that could integrate more readily into everyday apparel if challenges in actuator sizing and geometry are overcome.
\end{abstract}


\section{Introduction}
    % Exoskeletons are good.
    \IEEEPARstart{P}{owered} exoskeletons have long captivated our imagination, promising to enable humans to exceed their physical limitations. To this end, research into lower-limb exoskeletons has demonstrated substantial mobility improvements, bringing these devices closer to reality. Powered exoskeletons function by synchronously adding or removing mechanical energy alongside the human neuromuscular system. Targeted assistance applied using these devices has shown promising physiological benefits. For example, exoskeletons have improved endurance by reducing caloric demands and muscular effort required during activities including level and incline walking  \cite{mooney.herr:2014a, seo.shim:2016, seo.park:2017, kim.walsh:2019, lim.shim:2019, sawicki.young:2020, slade.collins:2022, luo.su:2024, bajpai.mazumdar:2024}, running \cite{kim.walsh:2018, kim.walsh:2019, sawicki.young:2020, lim.lee:2023, luo.su:2024}, climbing stairs \cite{kim.kim:2018, woo.rha:2021, luo.su:2024}, and carrying loads \cite{mooney.herr:2014, lee.walsh:2018, maclean.ferris:2019}, among others \cite{bywater.rouse:2023}. In addition, these devices may promote movement and independent living among elderly \cite{kim.kim:2018, jayaraman.jayaraman:2022} and individuals with mobility impairments \cite{rodriguez-fernandez.font-llagunes:2021} by enhancing their strength, balance, and endurance. Moreover, recent end-to-end exoskeleton control strategies have shown the ability to generalize assistance to new tasks, highlighting the accelerating pace of exoskeleton development \cite{molinaro.young:2024}. Consequently, exoskeletons have the potential to make a positive impact in the real-world by enhancing quality of life for older adults and individuals with mobility impairments, and supporting physically demanding activities in both recreational and occupational settings. Despite their growing prominence and encouraging results, the integration of these technologies into daily life remains limited. Modern exoskeleton hardware design strategies may result in exoskeleton physical layouts that extend beyond the body in indesirable ways, which can interfere with daily activities while affecting the aesthetics of the wearer \cite{yandell.zelik:2019}. New hardware design perspectives that promote form-fitting exoskeletons can help overcome these challenges.
    
    \begin{figure}[!t]
        \centering
        \includegraphics[width=\columnwidth]{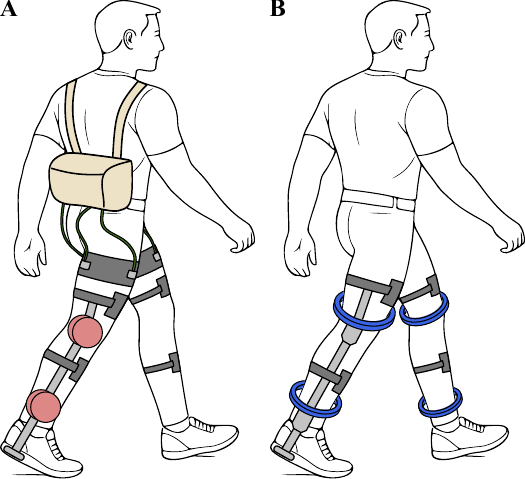}
        \caption{A: A traditional exoskeleton layouts with hip actuators placed remotely in a waist pack (yellow), while knee and ankle actuators (red) are positioned laterally in the sagittal plane. B: Limb-encircling layout, where the actuators (blue) encircle the assisted limb in the plane orthogonal to the limb axis.}
        \label{fig:form_factor}
        \vspace{-5mm}
    \end{figure}

    \begin{figure*}[!t]
        \centering
        \includegraphics[width=\textwidth]{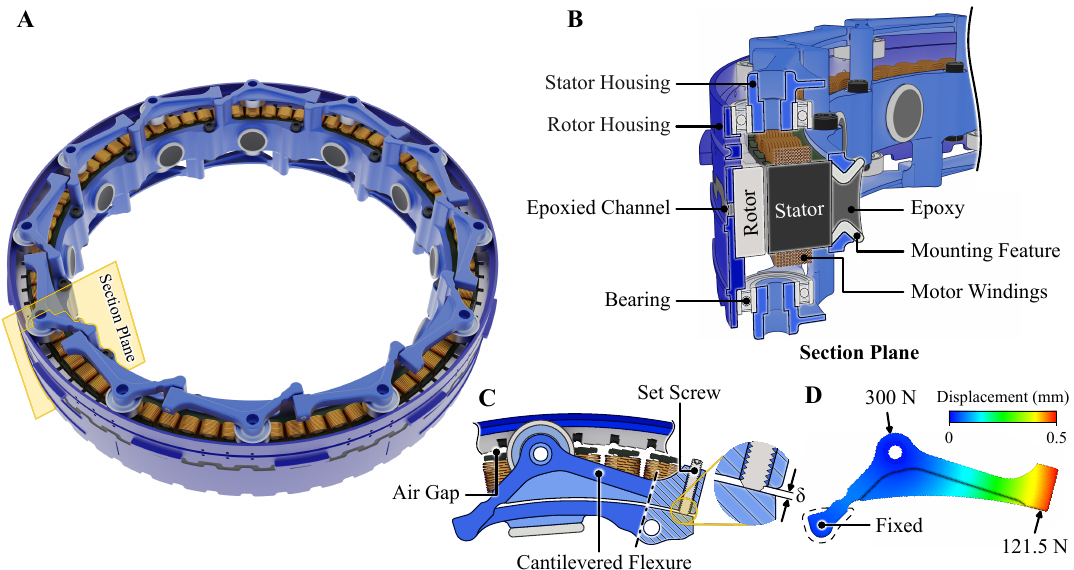}
        \caption{A: Rendering of the Limb-Encircling Actuator (LEA). B: Cross-sectional view illustrating the frameless motor and the housing structure comprising the LEA. The housing structure encloses the motor axially, with the rotor bonded to the rotor housing (dark blue) via an epoxy-filled channel, and the stator secured to the stator housing (light blue) with radially distributed mounting features (light grey) permanently bonded to the stator. C: Enlarged view of a cantilevered flexure with a mounted radial bearing. Set screws inserted in the flexures locally adjust the air gap between the rotor and stator by preloading the bearings. The distance $\delta$ defines the three bearing preload conditions -- \textit{unloaded}, \textit{design}, and \textit{loaded} -- for characterization experiments. D: Finite Element Analysis (FEA) of the cantilevered flexure was used to determine its allowable deflection range based on the static radial load capacity of the bearings.}
        \label{fig:actuator}
        \vspace{-5mm}
    \end{figure*}
    
    % \IEEEpubidadjcol   
    
    % The size and shape of an exoskeleton govern its experience, and the actuators within an exoskeleton drive its design.
    The physical characteristics and layout of the actuation subsystem within an exoskeleton govern its size, shape, and fit. Mechanical assistance is often provided by brushless electric motors with an integrated transmission, packaged together as an \textit{actuator} \cite{yu.su:2020, zhu.gregg:2021, qian.fu:2023}. Exoskeleton actuators are typically oriented in the sagittal plane \cite{rodriguez-fernandez.font-llagunes:2021, bajpai.mazumdar:2024, leestma.young:2024, zhao.gregg:2024} or are remotely located to the waist or back \cite{lee.walsh:2018, kim.walsh:2018, sanchez-villamanan.pons:2019, zhong.zhang:2022} (Fig.~\ref{fig:form_factor}A). Exoskeletons providing considerable assistive effort often have large actuators that weigh many hundreds of grams. Most often, actuators are positioned on the lateral aspect of the leg in the sagittal plane, with their rotation axis parallel or collinear with the axis of the assisted joint. Aligning the actuator and joint axis enables simpler designs that eliminate the need for linkages or bulky transmissions typically required for remote actuation. The vast majority of exoskeletons are developed using this design approach and have achieved promising outcomes across different populations \cite{seo.park:2017, kim.kim:2018, maclean.ferris:2019, bajpai.mazumdar:2024}. In addition, this actuation approach leverages commercial, off-the-shelf actuation options, which are tightly packaged and available across relevant power regimes \cite{leestma.young:2024, molinaro.young:2024, zhao.gregg:2024, nesler.gregg:2022}. However, a common challenge is the wider profile necessitated by the actuator's protrusion beyond the human body's physical envelope. Some exoskeletons can extend the lateral profile of the legs between 41-125 mm \cite{zhao.gregg:2024, nesler.gregg:2022, yu.su:2020, ishmael.lenzi:2022, qian.fu:2023, cao.wu:2023, leestma.young:2024}, which can impede movement in the community due to inadvertent impacts between the device and the objects in the environment. Another design strategy eliminates these lateral protrusions by remotely locating the actuators in wearable packs at the waist or back. However, these packs can interfere with essential activities and add to the physical burden on the wearer \cite{asbeck.walsh:2013, yandell.zelik:2017, kim.walsh:2018, siviy.walsh:2023}.  

    % Limb-encircling layout overview
    One strategy for developing more form-fitting exoskeleton designs is to explore how actuation mass is distributed relative to the human body. To this end, we propose redistributing the actuator mass and structure circumferentially around the limb in the transverse plane such that the actuator and limb axes are collinear (Fig.~\ref{fig:form_factor}A). In this limb-encircling layout, the actuator is worn over the limb by inserting the distal limb segments into the actuator cavity and sliding it to its resting position, similar to donning a sock or pants. Reorienting the actuator redistributes its required mass and volume uniformly around the limb rather than concentrating it along the lateral aspect of the limb. A limb-encircling actuation layout could support the development of exoskeletons that are more form-fitting or more easily integrated into apparel or existing, application-driven systems (e.g. space suit).  To our knowledge, there are no explorations of this actuation and application co-objective. 

    % 7. Contributions
    In this work, we introduce a limb-encircling layout able to distribute actuator mass and structure around the human body. This \textit{Limb-Encircling Actuator (LEA)} is based on a custom, frameless brushless motor that was designed for use in a unilateral ankle exoskeleton. The LEA design includes lightweight structural components that enable the integration of the frameless motor with the transmission. The LEA also features a novel layout of radial bearings that reduces the overall mass and cost of the actuator. This bearing assembly incorporates a preloading mechanism that maintains the air gap between the frameless motor's rotor and stator. We further describe the design of the LEA (Sec. \ref{design}) and characterize its electromechanical properties for multiple preload configurations of the bearing assembly using a modular testbed. We also evaluated the implications of motor size on exoskeleton performance by comparing the energy consumption of the LEA with brushless motors commonly used in exoskeletons across multiple activities of daily living (Sec. \ref{methods}, \ref{results}).

\section{Design} \label{design}
    \subsection{Limb-Encircling Actuator}
        % Actuator sizing & design requirements
        \noindent The design of the \textit{limb-encircling actuator} (LEA) is driven by the physical constraints of distributing the actuator structure around the limb. To enable donning and doffing the actuator over the assisted limb, the internal diameter of the actuator must be larger than the widest cross-section of the limb distal to the one about which it is worn. At the same time, its outer diameter must be minimized to reduce mass and avoid interference with adjacent limbs and the environment. Additionally, the actuator must facilitate the integration of transmission geometry which transfers motion to the orthogonal joint axis. That is, by locating the actuator in the transverse plane, its action must be rotated 90$^\circ$ by the transmission to apply mechanical power to the joint axis. To our knowledge, no commercially available actuators satisfy these design requirements. Hence, we developed a custom frameless motor and accompanying actuator structure, which together comprise the LEA. 
        
        % Frameless motor development
        We developed a custom frameless motor by specifying its geometric and electromechanical properties for integration in an ankle exoskeleton. For this application, we chose an exterior rotor motor because they provide greater torque capacity relative to an interior rotor configuration \cite{sensinger.schorsch:2011}. We specified the stator inner diameter (\textgreater145~mm) to fit over the largest aspect of the foot-ankle complex for 99th percentile of the US population \cite{fryar.ogden:2021}. We further specified the frameless motor to have a low rotor-stator thickness (\textless20~mm), and small axial length (\textless13~mm) to achieve a thin profile and minimize weight and bulk around the limbs. Other physical and electromechanical characteristics of the motor such as the torque constant were extrapolated from properties of a commercially available brushless motor (Model: U8-KV100; Manufacturer: T-motor, Nanchang, Jiangxi, China) \cite{lee.rouse:2019} as a function of air-gap radius \cite{seok.kim:2012, urs.moore:2022}. In this analysis, rotor-stator thickness and material properties of the custom frameless and commercial motors were assumed to be identical. In addition, we specified the peak current and voltage requirements  (\textgreater30~A, \textless24~V) such that the actuator could provide $\sim$50\% of the biological ankle torque during walking for a 70 kg individual and support ankle velocities necessary for running \cite{reznick.gregg:2021}. Based on these design criteria and estimated characteristics, the custom frameless motor was designed and manufactured by a commercial motor manufacturer (T-motor, Nanchang, Jiangxi, China). The resulting frameless motor had inner and outer diameters of 150 and 180~mm, respectively, an axial length of 13~mm, and weighed 600~g.
    
        % Frame design
        The design of the LEA includes structural components that house, orient, and position the frameless motor in the transverse plane. Additional structures were necessary to integrate the frameless motor within the exoskeleton design. The rotor and stator are housed within structural components machined from Aluminum alloy (7075-T6) \mbox{(Fig. \ref{fig:actuator}A)}, which is paramagnetic and minimizes interference with the motor's magnetic field. The rotor and stator are each enclosed between two structural parts \mbox{(Fig. \ref{fig:actuator}B)}, which mate axially, securing the rotor and stator between. The stator housing encloses the stator by capturing radially-spaced mounting features, which are permanently epoxied (Model: DP420NS; Manufacturer: 3M, Maplewood, Minnesota, USA) to the inner aspect of the stator. We chose this attachment method for the stator housing because of its low profile and mass, as well as the ease of disassembling/reassembling. The rotor housing is permanently secured to the rotor with an epoxy-filled annular channel (see Fig. \ref{fig:actuator}B). In addition, the housing structures incorporate a cable-routing channel and attachment points for interfacing with a cable-driven transmission. Relative motion between the rotor and stator is achieved using radially-spaced bearings mounted on cantilevered flexures integrated into the stator structure \mbox{(Fig. \ref{fig:actuator}C)}. The resulting actuator \mbox{(motor + framing)} weighs 894~g and has a nominal and 20-second peak torque capacities of 7.0~Nm and 16.3~Nm, respectively. 

        \begin{figure*}[!t]
            \centering
            \includegraphics[width=\textwidth]{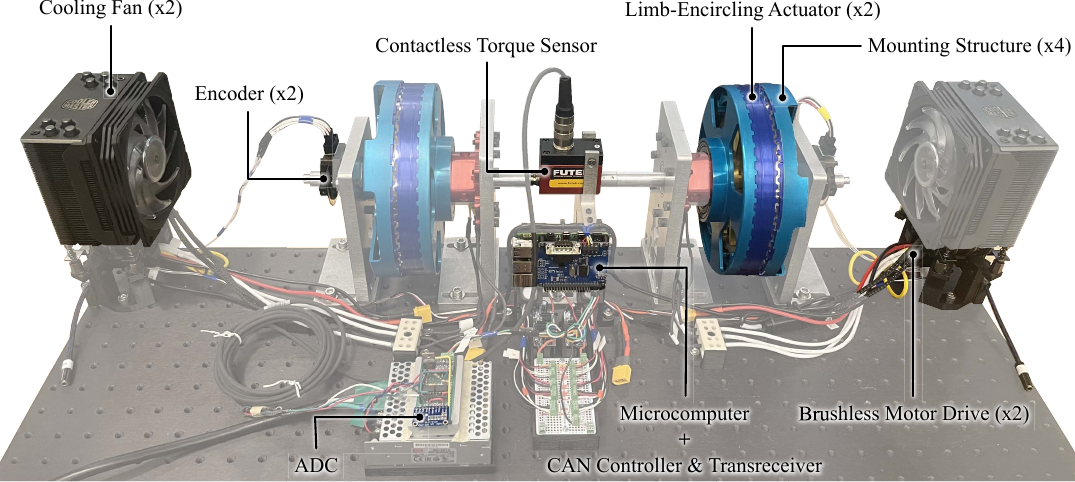}
            \caption{Custom-built testbed developed to characterize the Limb-Encircling Actuator (LEA). The LEAs (blue) are enclosed within structural members (teal) and mounted on shafts connected at their opposite ends through a contactless torque sensor. Brushless motor drives commutate the actuators, with angular position feedback provided by encoders installed at the shaft ends. A single-board computer equipped with a Controller Area Network (CAN) controller and transceiver sends control trajectories to the motor drives, and performs data collection from an ADC connected to the load cell. Cooling fans were added to dissipate heat from the motor drives, although their cooling capacity exceeds the demands of this application.}
            \label{fig:testbed}
            \vspace{-5mm}
        \end{figure*}
        
        % Bearing of bearings
        The LEA features a unique bearing layout driven by the lower mass and cost. The rotor and stator of electric motors are separated by an air gap that allows their relative motion. This air gap is typically maintained by constraining the rotor and stator using pairs of radial bearings. However, bearings suitable for the large diameter of the LEA substantially increase its weight and cost. Instead, we integrated a circular array of 12 miniature radial ball bearings on either side of the LEA \mbox{(Fig. \ref{fig:actuator}A)}, termed the \textit{bearing of bearings}. Each of these bearings is mounted on a cantilevered flexure \mbox{(Fig. \ref{fig:actuator}C)}. The preload between each bearing and the stator can be adjusted via tensioning set screws within the flexures, allowing local adjustment of the air gap between the stator and rotor. Preloading the cantilevered flexures helped address any physical deformation of the LEA caused by magnetic forces between the rotor and stator, and compensate for dimensional deviations from manufacturing tolerances. The free end of the flexure can accommodate a deflection of 0.5~mm before the bearings reach their maximum static radial load capacity of 300~N \mbox{(Fig. \ref{fig:actuator}D)}.

    \subsection{Actuator Characterization Testbed}    
        \noindent We built a custom testbed to evaluate the LEA design and characterize its electromechanical properties. The testbed \mbox{(Fig. \ref{fig:testbed})} was comprised of two actuators that were enclosed in structural members and mounted on aluminum shafts. The shafts were connected to either end of a contactless torque sensor (Model: TRD605 FSH02025; Manufacturer: Futek Advanced Sensor Technology, Inc., Irvine, CA, USA). Each actuator was controlled independently via a brushless motor drive (Model: G-SOLTWIR50/100 SE1S; Manufacturer: Elmo Motion Control, Petach-Tikva, Israel) using field-oriented control commutation. Two lithium polymer batteries were connected in series to power the brushless motor drives, supplying a nominal voltage of 29.6~V (Model: 4S 7000 mAH; Manufacturer: Venom Power, Kowloon, Hong Kong). Heat sinks with cooling fans were added to dissipate the heat produced by the brushless drives, though their cooling capacity exceeded the requirements of this application. Position feedback was provided to the brushless drives using optical encoders (Model: EM2, \mbox{10000 lines/rev}; Manufacturer: US Digital, Vancouver, WA, USA) mounted on either of the shaft ends. Onboard drive sensing provided q-axis current, shaft angle, and velocity measurements. Control trajectories were commanded, and sensor data was collected using either a single-board computer (Model: Raspberry Pi 4B; Manufacturer: Raspberry Pi Foundation, Cambridge, UK) or a laptop PC. The single-board computer communicated with the brushless drives via Controller Area Network (CAN), utilizing an additional board (Model: PiCAN2; Manufacturer: Copperhill Technologies, Greenfield, MA, USA) and a custom Python driver. Communication between the drives and the laptop PC was performed using the manufacturer’s proprietary software (Elmo Application Studio) over an RS-232 connection. The testbed also included a 16-bit ADC (Model: ADS1115; Manufacturer: Texas Instruments, Dallas, TX, USA) for measuring actuator torque via the contactless torque sensor.
        
\section{Methods} \label{methods}
    \subsection{Actuator Dynamics}
        \noindent We represent the dynamics of the LEA using the brushed DC model to simplify its analysis and control \cite{lee.rouse:2023}. This representation utilizes the power-invariant direct-quadrature transformation \cite{park:1929, lee.rouse:2023} to convert the sinusoidally varying winding currents and voltages into their DC equivalents while accounting for the motor winding type. The electrical dynamics of the actuator are modeled using Kirchhoff's Voltage Law as,
        \begin{equation}
            V^q = R^\phi I^q + L^e \frac{dI^q}{dt} +  K_b^q \dot{\theta}_r
            \label{eq:KVL}
        \end{equation}
        where, $V^q$ is the q-axis voltage, $R^\phi$ is the phase resistance, $I^q$ is the q-axis current, $L^e$ is the effective inductance, $K_b^q$ is the q-axis back-EMF constant, and $\dot{\theta}_r$ is the angular velocity of the rotor. The mechanical dynamics are modeled using Newton's second law as,
        \begin{equation}
            J \ddot{\theta}_r = K_t^q I^q - b \dot{\theta_r} - f\,\text{sgn}(\dot{\theta}_r) - \tau_L
            \label{eq:newtons_2nd_law}
        \end{equation}
        where, $J$ is the rotational moment of inertia, $\ddot{\theta}_r$ is the angular acceleration of the rotor, $K_t^q$ is the q-axis torque constant, $b$ is the coefficient of viscous damping, $f$ is the coulomb friction, and $\tau_L$ is the load torque. 
    
    \subsection{Bearing Preload Configurations} \label{preload_config}
        We characterized the LEA under different bearing preload configurations to evaluate their impact on actuator performance. While a minimum preload is necessary to maintain the air gap between the rotor and stator, excessive preload increases coulomb friction and viscous damping, thereby reducing actuator efficiency. To quantify this effect, we evaluated the LEA under three distinct preload conditions: \textit{unloaded}, \textit{design}, and \textit{loaded}. The bearing preloads were adjusted with tensioning set screws by modifying the gap ($\delta$) between the cantilevered flexures and the stator housing \mbox{(Fig. \ref{fig:actuator}C)}. In the \textit{unloaded} configuration, the flexures remained in their manufactured condition, with set screws contacting the stator to prevent deflection, corresponding to zero preload. However, actual part dimensions may deviate from nominal values due to manufacturing tolerances. As a result, the gap ($\delta$) in the assembled actuator was smaller than the intended value for some flexures. In the \textit{design} configuration, $\delta$ was increased to 0.5~mm to match the target design gap. In the \textit{loaded} configuration, $\delta$ was further increased to 0.75~mm, corresponding to an expected preload of 138~N, as estimated via finite element analysis. 

    \subsection{Statistics and Comparisons}
        We applied a one-way analysis of variance (ANOVA) to evaluate the statistical significance of the actuator's estimated parameters. In this analysis, the bearing preload configurations (\textit{unloaded}, \textit{design}, and \textit{loaded}) were treated as the fixed factors, and the actuator was considered a random factor. The dependent variables consisted of the actuator parameters, including the q-axis torque constant, coulomb friction, and viscous damping. For significant main effects, post-hoc comparisons were performed using Tukey's HSD with a significance level of 0.05 \cite{mickey.clark:2009}.

    \subsection{System Identification}
        \noindent The electromechanical properties of the LEA were characterized across three different bearing preload configurations -- \textit{unloaded}, \textit{design}, and \textit{loaded} \mbox{(Sec. \ref{preload_config})}. The characterization experiments were performed using a single-board computer, which commanded control trajectories and collected data at approximately 50~Hz unless specified otherwise.
 
        \subsubsection{Controller Performance}
            The performance of the closed-loop current and position controllers onboard the brushless drives were evaluated in the time and frequency domains. In these tests, the actuator was mechanically grounded in current control and allowed to spin freely in position control. Data was collected using the manufacturer's proprietary software at 1000~Hz, while reference trajectories were provided by the single-board computer at average rate of 900~Hz for current control and 300~Hz for position control. Time domain performance was evaluated using the step responses with magnitudes of 2.5~A, 5.0~A, and 7.5~A for current control, and $\pi/4$, $\pi/2$, and $3\pi/4$ for position control. Five trials were performed for each step magnitude, with each trial lasting 2~s. Rise time, overshoot, and settling time were computed for each trial and averaged across trials and reference levels. Frequency domain performance was evaluated using responses to randomly sampled Gaussian reference trajectories with means matched to the step input magnitudes. These trajectories were low-pass filtered using a Butterworth filter with cut-off frequencies of 218~Hz and 40~Hz, and filter orders of 35 and 10, for current and position control, respectively. The recorded data was then filtered using low-pass Butterworth filter with orders of 10 and 5, and cut-off frequencies of 225~Hz and 50~Hz for the corresponding signals. The frequency response was determined using Blackman-Tukey spectral analysis \cite{ljung:1998} with a window size of 400 data points. The controller bandwidths were determined at the -3~dB crossover point and averaged across multiple trials and reference levels. Five trials were conducted for each controller. Each current control trial comprised five 8.192~s long sub-trials to increase the frequency resolution, whereas each position control trial lasted 8.192~s only.

        \subsubsection{Torque Constant}
            The torque constant is a fundamental motor property that governs the relationship between the winding current and applied torque. For this experiment, one actuator applied torque using current control while the other maintained a fixed position using position control. Current references were commanded, and the corresponding torques were measured under different bearing preloads. For each trial, currents ranging from 4.9~A to 24.5~A were applied in 4.9~A increments. Data were recorded for \mbox{5.5 s} for each current setpoint, with the first \mbox{0.5 s} discarded to allow the system to reach a steady state. Linear regression was performed on the averaged currents and torques using, 
            \begin{equation}
                K_t I^q = \slashed{J\ddot{\theta}_r} + \slashed{b\dot{\theta}_r} + \slashed{f\,\text{sgn}(\dot{\theta}_r)} + \tau_L,
                \label{eq:torque_constant}
            \end{equation}
            where inertial and damping terms were omitted because the setup was stationary, and the frictional effects were assumed to be negligible. Six trials were performed, and the average torque constant across all the trials was reported. The goodness of fit for the linear regression was evaluated using the coefficient of determination ($R^2$).

        \subsubsection{Back-EMF Constant}
            The back-EMF constant defines the relationship between the rotor's angular velocity and the voltage induced across the windings. To determine the LEA's back-EMF constant, line-to-line winding voltages were measured as the actuator was externally driven at 2$\pi$~rad/s. Voltages were recorded with a data acquisition board (Model: USB-6218; Manufacturer: National Instruments, Austin, TX, USA) at 10~kHz for five seconds. The sinusoidal signals were segmented into individual waves to extract signal amplitude and frequency. The amplitude of the line-to-line back-EMF constant was calculated using

            \begin{equation}
                V^{ll}(\theta) = K_b^{ll}(\theta)\ \dot{\theta}_r
            \end{equation}
            
            where $V^{ll}$ is the line-to-line voltage, $K_b^{ll}$ is the line-to-line back-EMF constant, and $\theta$ is the magnetic angle. The rotor's angular velocity, $\dot{\theta}_r$, was determined using

            \begin{equation}
                \dot{\theta}_r = \frac{2\pi}{n_p} f^{ll}
            \end{equation}

            where, $n_p$ is the number of pole pairs, and $f^{ll}$ is the frequency of the back-EMF voltage. The line-to-line back-EMF constant was converted to its q-axis equivalent ($K^q_b$) using the appropriate conversion for the motor's winding configuration (Wye) \cite{lee.rouse:2023}:

            \begin{equation}
                K_b^q = \frac{1}{\sqrt{2}} K_b^{ll}
            \end{equation}
        
        \subsubsection{Friction and Damping}
            Coulomb friction and viscous damping contribute to the mechanical inefficiencies in an actuator. These parameters were estimated by applying controlled currents and measuring the no-load actuator speeds under different bearing preloads on an actuator decoupled from the testbed. Each trial involved applying positive and negative currents in 0.03~A increments with magnitudes ranging from 0.6~A - 0.74~A for the \textit{unloaded} and \textit{design} configurations, and from 0.76~A - 0.89~A for the \textit{loaded} configuration. Data collection spanned 10 s per current reference, with the first five seconds excluded from the analysis to allow the actuator to achieve steady state conditions. The steady-state currents and velocities were averaged for each current reference and used to estimate the parameters via linear regression. Separate linear regression was performed on the positive and negative current and velocity pairs using,
            \begin{equation}
                K_t I^q = \slashed{J\ddot{\theta}_r} + b\dot{\theta}_r + f\,\text{sgn}(\dot{\theta}_r) + \slashed[0][2pt][1.2]{\tau_L},
                \label{eq:friction&damping} 
            \end{equation}
            where inertial and load torque were discarded due to steady-state and no-load conditions. Five trials were conducted, and the average coulomb friction and viscous damping across all trials were reported. The goodness of fit for the linear regression was evaluated using the coefficient of determination ($R^2$).
     
        \subsubsection{Efficiency}
            The conversion of electrical energy to mechanical energy, or vice versa, by electrical motors is inherently limited by thermal and mechanical losses. The overall efficiency of an electric motor is governed by its operating torque-speed regime, and was empirically determined for a range of torque-speed combinations under different bearing preloads. In this experiment,  one actuator provided positive power, driving the motion, while the other actuator opposed the motion, operating in negative power. The driving actuator was controlled in velocity mode, with speeds ranging from 2.62~rad/s to 31.42~rad/s in 2.62~rad/s increments. Simultaneously, the driven actuator applied opposing torques using current control, with currents ranging from 1.2~A to 24~A in 1.2~A increments. Each trial lasted 3.5 s, and the initial 1.5~s were discarded to allow the system to reach steady state. The actuator efficiency was computed using the mean current and velocity for each trial as follows:
            \begin{align}
                \text{Driving:} \ &\eta_+ = \frac{\tau_L \dot{\theta}_r}{V^q_+ I^q_+} \\
                \text{Driven:} \ &\eta_{-}=\frac{V_- I^q_-}{\tau_L \dot{\theta}_r},
            \end{align}
            where $\eta$ is the efficiency, and the subscripts $+$ and $-$ denote whether the actuator is operating in positive power (driving) or negative (driven) mechanical power.

            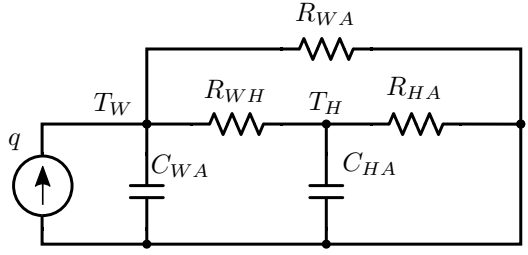
\begin{figure}[!t]
                \centering
                \def\svgwidth{.8\columnwidth}
                %% Creator: Inkscape 1.1.2 (b8e25be833, 2022-02-05), www.inkscape.org
%% Accompanies image file 'thermal_model_japman.pdf' (pdf, eps, ps)
\begingroup%
  \makeatletter%
  \providecommand\color[2][]{%
    \errmessage{(Inkscape) Color is used for the text in Inkscape, but the package 'color.sty' is not loaded}%
    \renewcommand\color[2][]{}%
  }%
  \providecommand\transparent[1]{%
    \errmessage{(Inkscape) Transparency is used (non-zero) for the text in Inkscape, but the package 'transparent.sty' is not loaded}%
    \renewcommand\transparent[1]{}%
  }%
  \providecommand\rotatebox[2]{#2}%
  \newcommand*\fsize{\dimexpr\f@size pt\relax}%
  \newcommand*\lineheight[1]{\fontsize{\fsize}{#1\fsize}\selectfont}%
  \ifx\svgwidth\undefined%
    \setlength{\unitlength}{141.07922745bp}%
    \ifx\svgscale\undefined%
      \relax%
    \else%
      \setlength{\unitlength}{\unitlength * \real{\svgscale}}%
    \fi%
  \else%
    \setlength{\unitlength}{\svgwidth}%
  \fi%
  \global\let\svgwidth\undefined%
  \global\let\svgscale\undefined%
  \makeatother%
  \begin{picture}(1,0.47458859)%
    \lineheight{1}%
    \setlength\tabcolsep{0pt}%
    \put(0,0){\includegraphics[width=\unitlength,page=1]{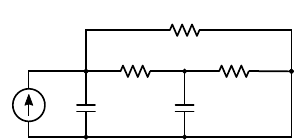}}%
    \put(0.62639489,0.42431857){\makebox(0,0)[t]{\lineheight{1.25}\smash{\begin{tabular}[t]{c}$R_{WA}$\end{tabular}}}}%
    \put(0.79266165,0.28621409){\makebox(0,0)[t]{\lineheight{1.25}\smash{\begin{tabular}[t]{c}$R_{HA}$\end{tabular}}}}%
    \put(0.45850101,0.28060482){\makebox(0,0)[t]{\lineheight{1.25}\smash{\begin{tabular}[t]{c}$R_{WH}$\end{tabular}}}}%
    \put(0.70972685,0.14419388){\makebox(0,0)[t]{\lineheight{1.25}\smash{\begin{tabular}[t]{c}$C_{HA}$\end{tabular}}}}%
    \put(0.35513545,0.14113908){\makebox(0,0)[t]{\lineheight{1.25}\smash{\begin{tabular}[t]{c}$C_{WA}$\end{tabular}}}}%
    \put(0.62869432,0.25743682){\makebox(0,0)[t]{\lineheight{1.25}\smash{\begin{tabular}[t]{c}$T_{H}$\end{tabular}}}}%
    \put(0.23047583,0.25731979){\makebox(0,0)[t]{\lineheight{1.25}\smash{\begin{tabular}[t]{c}$T_{W}$\end{tabular}}}}%
    \put(0.04553609,0.19382826){\makebox(0,0)[t]{\lineheight{1.25}\smash{\begin{tabular}[t]{c}$q$\end{tabular}}}}%
  \end{picture}%
\endgroup%

                \caption{Thermal model of the actuator represented as an equivalent electrical circuit.}
                \label{fig:thermal_model}
                \vspace{-5mm}
            \end{figure}
            
        \subsubsection{Thermal Dynamics}  
            The performance of an electric motor is limited by the Joule heating of its electrical windings. This heat is primarily dissipated through conduction to the actuator housing and convection to the surrounding air. These thermal dynamics can be modeled using an equivalent electrical circuit \mbox{(Fig. \ref{fig:thermal_model})}, represented as:
            \begin{equation}
                \renewcommand{\arraystretch}{1.5}
                \begin{bmatrix}
                     \dot{T_\text{W}} \\
                     \dot{T_\text{H}} 
                \end{bmatrix}
                =  \begin{bmatrix}
                     -\frac{R_{\text{WA}} + R_{\text{WH}}} {R_{\text{WA}} R_{\text{WH}} C_{\text{WA}}}  
                     & \frac{1}{R_{\text{WH}} C_{\text{WA}}} \\
                     \frac{1}{R_{\text{WH}} C_{\text{HA}}} & 
                     -\frac{R_{\text{HA}} + R_{\text{WH}}}{R_{\text{HA}} R_{\text{WH}} C_{\text{HA}}}
                \end{bmatrix} 
                \begin{bmatrix}
                     T_\text{W} \\
                     T_\text{H} 
                \end{bmatrix}
                + \begin{bmatrix}
                     \frac{1}{C_{\text{WA}}} \\
                     0
                \end{bmatrix} Q,
                \label{eq:thermal_dynamics}
            \end{equation}
            where $T$ is the temperature, $R$ is the thermal resistance, $C$ is the thermal capacitance, and $Q$ is the heat flux. The subscripts $W$, $H$, and $A$ represent the thermal properties of the motor windings, actuator housing, and ambient air, respectively.
                           
            The parameters of the thermal model (Fig.~\ref{fig:thermal_model}) were empirically estimated by electrically energizing the electrical windings and recording the actuator temperatures. A continuous current of 9~A was supplied to two electrical leads using a power supply (Model: 1688B; Manufacturer: BK Precision, Yorba Linda, CA, USA). Electrical power was supplied only to two leads to prevent the motor from spinning. Power was applied for approximately 75 minutes until the actuator temperatures reached a steady state, after which power was removed to allow the actuator to cool to room temperature. line-to-line voltages were recorded at 30~Hz using a data acquisition board (Model: USB-6218; Manufacturer: National Instruments, Austin, TX, USA). The heat flux ($Q$) from the windings was computed as the product of the supplied current and the recorded line-to-line voltages. Temperatures of the active and inactive windings, as well as the housing structure, were measured at three locations using a thermal camera (Model: 616C; Manufacturer: Fotric, Santa Clara, CA, USA) operating at 30 Hz. The mean housing temperature ($T_H$) was calculated by averaging the values across all locations. An equivalent winding ($T_W$) temperature was estimated by computing a weighted average of the mean temperatures of active and inactive windings across all locations, assuming uniform heat flux across all windings. The model parameters (eq. \ref{eq:thermal_dynamics}) were estimated using a nonlinear optimization algorithm \cite{fmincon}, which minimized the sum of the squared errors between the measured and predicted actuator temperatures ($T_H$, $T_W$). In addition, the temperature dependence of the line-to-line winding resistance ($R^{ll}$) was determined by fitting a linear model, 
            \begin{equation}
                R^{ll} = R^{ll}_o (1 + \alpha \Delta T),
                \label{eq:resistance_vs_t}
            \end{equation}
            where $R^{ll}_o$ is the initial resistance, $\alpha$ is the temperature coefficient of resistance, and $\Delta T$ is the difference between final and initial winding temperatures. 
            The estimated parameters of the thermal model, along with the winding resistance were used to predict the continuous and peak current (20~s) capacity of the LEA. 
            
            \begin{figure}[!t]
                \centering
                \includegraphics[width=\linewidth]{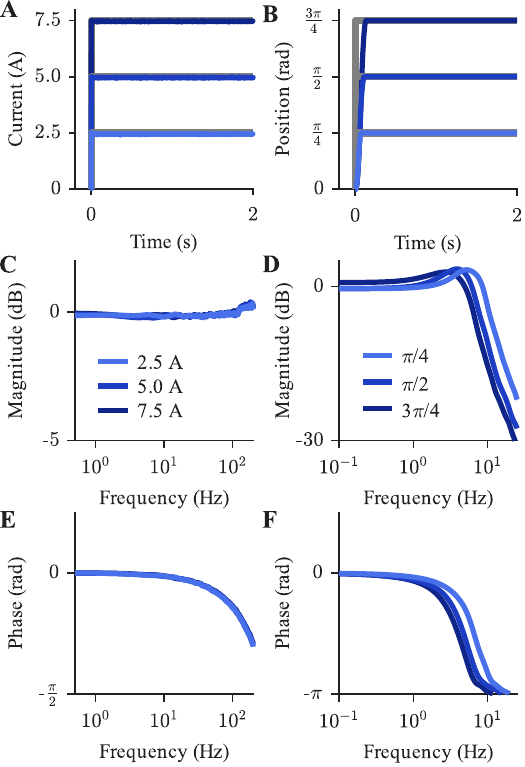}
                \caption{A: Current step responses to different q-axis current references. B: Position step responses to different reference positions. C: Current magnitude response to a white-noise input with a 200 Hz cutoff. D: Position magnitude response to a white-noise input with 25 Hz cutoff. E: Current phase response to a white-noise input with a 200 Hz cutoff. F: Position phase response to a white-noise input with a 25 Hz cutoff.}
                \label{fig:control_perf}
            \vspace{-5mm}
            \end{figure}
            
            \begin{figure*}[!t]
                \centering
                \includegraphics[width=\textwidth]{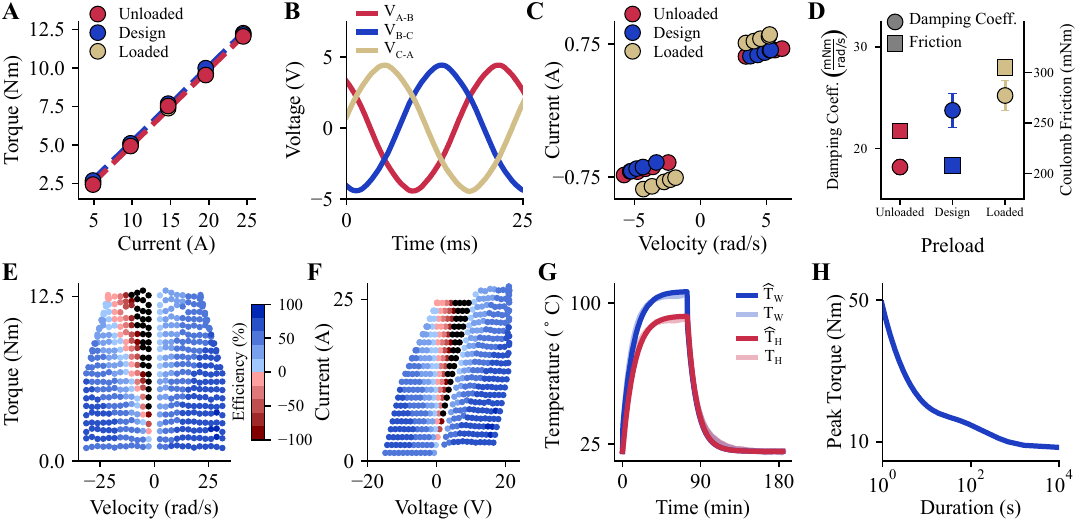}
                \caption{A: Torque and current relationship across different bearing preloads. The torque constant was determined from the slope of best-fit lines. Error bars for torques represent one standard deviation, but are not visible due to their small magnitude. B: Sinusoidal line-to-line back-EMF voltages measured across the three winding pairs. C: No-load speeds of the actuator across different bearing preloads. Error bars represent one standard deviation of velocities, but are not visible due to their small magnitude. D: Estimated coefficient of viscous damping and coulomb friction for each preload configuration, with error bars indicating one standard deviation. The coefficients of viscous damping and coulomb friction were estimated from the slope and intercept of the best-fit lines, respectively. E: LEA efficiency across different torque and velocity combinations for the \textit{design} configuration. F: LEA efficiency across different current-voltage regimes for the \textit{design} configuration. Negative efficiencies were observed when the voltage drop across the phase resistance was higher in magnitude and opposite in sign than the back-EMF voltage. Black markers indicate efficiencies less than -100\%. G: Thermal response of the LEA showing the measured and estimated winding and housing temperatures. H: Peak actuator torque estimated for different operating durations using the thermal model.}
                \label{fig:sys_id}
                \vspace{-5mm}
            \end{figure*}
            
    \section{Results} \label{results}
        \noindent The results of the system identification experiments are presented for three different bearing preload configurations---\textit{unloaded}, \textit{design}, and \textit{ loaded}---of the LEA \mbox{(Sec. \ref{preload_config})}.
    
        \subsection{Controller Performance}
            \noindent The closed-loop current and position controllers demonstrated precise reference tracking \mbox{(Fig. \ref{fig:control_perf}A, \ref{fig:control_perf}B)} with sufficient bandwidths \mbox{(Fig. \ref{fig:control_perf}C, \ref{fig:control_perf}D}) to support human locomotion \cite{winter:2009}. The average rise time, settling time, and overshoot computed across all trials and reference levels were 1~ms, 8.3\%, and 90~ms, respectively for current control, and 57.8~ms, 0.1\%, and 91~ms respectively, for position control. Both current and position controllers maintained steady-state errors below 0.1\%. The average -3~dB crossover bandwidth determined across all trials and reference levels was \textgreater200~Hz for current control and 7.5~Hz for position control.
                
        \subsection{Torque Constant}
            \noindent The torque constant remained consistent across all three bearing preload configurations \mbox{(Fig. \ref{fig:sys_id}A)}, with an estimated value of \mbox{0.49$\pm$ 0.06 Nm/A}. The analysis of variance (ANOVA) indicated no significant effect of bearing preload on the torque constant (\mbox{$p$ = 0.98}, \mbox{$F_{2,15}$ = 0.02}). This invariance is likely due to the constant air gap between the rotor and stator maintained by the actuator’s mounting on the testbed. Additionally, unmodeled friction effects had minimal impact on the estimated torque constant, as indicated by the low standard deviation of the estimates. The actuator model strongly agreed with the experimental data, with $R^2$ values of the best fit lines exceeding 0.99.    
        
        \subsection{Back-EMF Constant}
            The LEA exhibited a \noindent sinusoidal back-EMF profile (Fig. \ref{fig:sys_id}B), with a back-EMF constant numerically equivalent to the torque constant. The back-EMF constant was estimated to be \mbox{0.49 $\pm$ 0.004} \small$\frac{\text{V}}{\text{rad/s}}$ \normalsize. The back-EMF constant was consistent across all three winding pairs, indicating uniform winding characteristics.
                
        \subsection{Friction and Viscous Damping}
            \noindent Viscous damping increased with bearing preload, while coulomb friction was lowest in the \textit{design} configuration \mbox{(Fig. \ref{fig:sys_id}C, \ref{fig:sys_id}D)}. The coefficients of viscous damping were \mbox{18.2 $\pm$ 9.7 \small$\frac{\text{mNm}}{\text{rad/s}}$}\normalsize, \mbox{23.8 $\pm$ 8.0 \small$\frac{\text{mNm}}{\text{rad/s}}$}\normalsize, and \mbox{25.5 $\pm$ 11.1 \small$\frac{\text{mNm}}{\text{rad/s}}$} \normalsize for \textit{unloaded}, \textit{design}, and \textit{loaded} configurations, respectively. Corresponding coulomb friction values were \mbox{241.9 $\pm$ 43.4 mNm}, \mbox{208.0 $\pm$ 38.4 mNm}, and \mbox{303.7 $\pm$ 73.5 mNm}. Both viscous damping (\mbox{$p$ = 6.5$\times$10$^{-5}$}, \mbox{$F_{2,27}$ = 14.09}) and coulomb friction (\mbox{$p$ = 1.4$\times$10$^{-11}$}, \mbox{$F_{2,27}$ = 72.47}) varied significantly with bearing preload. Post-hoc comparisons showed that the \textit{design} and \textit{loaded} configurations exhibited significantly higher damping ($p$ \textless\ 0.05) than the \textit{unloaded} configuration but had no significant differences between them. All configurations differed significantly in coulomb friction ($p$ \textless\ 0.05). Linear regression models strongly agreed with the data, as indicated by $R^2$ values of 0.97, 0.93, and 0.97 for the \textit{unloaded}, \textit{design}, and \textit{loaded} configurations, respectively.
        
        \subsection{Efficiency}
            \noindent Efficiency remained comparable but slightly decreased with increasing bearing preload. The efficiencies in the positive power region ($\tau_L$ \textgreater\ 0, $\dot{\theta}r$ \textgreater\ 0) were \mbox{53.7 $\pm$ 52.0\%}, \mbox{53.5 $\pm$ 52.2\%}, and \mbox{51.6 $\pm$ 46.4\%} for the \textit{unloaded}, \textit{design}, and \textit{loaded} configurations, respectively. Both positive and negative efficiencies were observed in the negative power region ($\tau_L$ \textgreater\ 0, $\dot{\theta}r$ \textless\ 0). Notably, negative efficiencies emerged in low-velocity and high-torque regimes ($\tau_L$ + 0.5$\dot{\theta}_r$ \textgreater\ 0), indicating that the thermal power loss was greater than the negative mechanical power input. The LEA was more efficient ($\eta_+$ \textgreater\ 75\%) at lower torques ($\tau_L$ \textless\ 6.0 Nm) and higher velocities ($\dot{\theta}_r$ \textgreater\ 23.5 rad/s) (Fig. \ref{fig:sys_id}E, \ref{fig:sys_id}F).
                  
        \subsection{Thermal Dynamics} 
            \noindent The thermal model predictions closely agreed with experimentally measured temperatures (Fig.~\ref{fig:sys_id}G). Nonlinear optimization identified the thermal model parameters with a root-mean-square error of 1.7~$^\circ$C. The estimated thermal resistances for winding-ambient ($R_{WA}$), housing-ambient ($R_{HA}$), and winding-housing ($R_{WH}$) paths were 5.6~K/W, 1.7~K/W, and 0.3~K/W, respectively. The identified thermal capacitances of the windings and housing were 22.9~\small$\frac{\text{W}}{\text{K/s}}$\normalsize,  469.1~\small$\frac{\text{W}}{\text{K/s}}$\normalsize, and 469.1\small$\frac{\text{W}}{\text{K/s}}$\normalsize, respectively. Linear regression estimated the line-to-line winding resistance ($R^{ll}$) as 528~m$\Omega$, and the coefficient of thermal resistance ($\alpha$) as 4$\times$10$^{-4}$~s$^{-1}$ with an R$^2$ value of 0.99. The thermal model was used to estimate a continuous current capacity of 13.7~A, and peak current as a function of motor operating time (Fig:~\ref{fig:sys_id}H), assuming a peak winding temperature of 130~$^\circ$C.

\section{Discussion} \label{discussion}
    \begin{table}[!t]
        \centering
        \caption{Electromechanical Specifications of the Limb-Encircling Actuator (LEA), U8-KV100 (T-Motor), EC-4Pole 30 (Maxon)}
        \label{tab:motor_specs}
        
        \setlength{\tabcolsep}{2pt}
        \renewcommand{\arraystretch}{1.1}
   
        \begin{tabular}{l c c c}
            \hline \hline
    
            \textbf{Dimensions} &
            \textbf{LEA} &
            \textbf{U8-KV100}\cite{lee.rouse:2019} &
            \textbf{EC-4P 30}\cite{maxon.ec4pole.30.200w} \\[6pt]
            
            Outer Diameter (mm) & 182 & 85 & 30\\
            Inner Diameter (mm) & 138 & 54 & --\\
            Axial Length (mm) & 40 & 27 & 64\\[2pt]
            
            \multicolumn{4}{l}{\textbf{Mechanical}} \\
            Mass (g) & 894 & 240 & 300\\ 
            Rotor Inertia, $J$ (g$\cdot$m$^2$) & 2.4 $^a$ & $12.0\times10^{-3}$ & $3.3\times10^{-3}$\\
            Damping Coefficient $\left( \frac{\text{mNm}}{\text{rad/s}} \right)$ & 23.8 $^b$ & 0.2 & $5.7\times10^{-3}$\\
            Coulomb Friction (mNm) & 208 $^b$ & 17 & --\\[2pt]
            
            \multicolumn{4}{l}{\textbf{Electrical}}\\
            Torque Constant, $k_t^q$ (Nm/A) & 0.49 & 0.14 & 0.02\\
            Motor Constant, $k_m^q$ (Nm/$\sqrt{\text{W}}$) & 0.95 & 0.27 & 0.04\\
            Number of Phases & 3 & 3 & 3\\
            Number of Pole Pairs & 41 & 21 & 2\\
            Terminal Resistance (m$\Omega$) & 528 & 186 & 210\\
            Terminal Inductance ($\mu$H) & 255.9 $^c$ & 138 & 37\\
            Winding Type & Wye & Delta & Wye\\[2pt]
    
            \multicolumn{4}{l}{\textbf{Control}}\\
            \multicolumn{4}{l}{Current:}\\
            $\qquad$Rise Time (ms) & 1.0 & 5.0 & -- \\
            $\qquad$Overshoot (\%) & 8.3 & 39.8 & -- \\
            $\qquad$Settling Time (ms) & 90.0 & 30.2 & -- \\
            $\qquad$Steady-state Error (\%) & 0.1 & 0.6 & -- \\
            $\qquad$Bandwidth (Hz) & \textgreater 200 & \textgreater 320 & -- \\
            \multicolumn{4}{l}{Position:}\\
            $\qquad$Rise Time (ms) & 57.8 & 22.2 & -- \\
            $\qquad$Overshoot (\%) & 0.1 & 8.45 & -- \\
            $\qquad$Settling Time (ms) & 91 & 60.3 & -- \\
            $\qquad$Steady-state Error (\%) & 0.1 & 0.4 & -- \\
            $\qquad$Bandwidth (Hz) & 7.5 & 18 & -- \\            
            
            \multicolumn{4}{l}{\textbf{Thermal}}\\
            \multicolumn{4}{l}{Resistance:}\\
            $\qquad$Winding-Ambient (K/W) & 5.6 & 3416.3 & --\\
            $\qquad$Winding-Housing (K/W)  & 0.3 & 1.1 & 0.2\\
            $\qquad$Housing-Ambient (K/W)  & 1.7 & 3.5 & 7.4\\
            \multicolumn{4}{l}{Capacitance:}\\
            $\qquad$Winding (s) & 22.9 & 25.8 & 10.0\\
            $\qquad$Housing (s) & 469.1 & 130.6 & 159.5\\
            Max. Continuous Current (A) & 13.7 & 7.6$^d$ & 5.1\\[2pt]
            
            \hline\hline
            \noalign{\vskip 2pt}
            \multicolumn{4}{l}{a: Measured from CAD} \\
            \multicolumn{4}{l}{b: For \textit{design} preload configuration} \\
            \multicolumn{4}{l}{c: Measured using LCR Meter} \\
            \multicolumn{4}{l}{$\quad $ (Model: U1733C; Manufacturer: Keysight Technologies, CA, US)} \\
            \multicolumn{4}{l}{d: Updated from Lee et al. \cite{lee.rouse:2019} to account for change in winding} \\
            \multicolumn{4}{l}{$\quad$resistance with temperature.}
        \end{tabular}
        \vspace{-5mm}
    \end{table}
    
    % What we did and why
    \noindent In this work, we introduced the limb-encircling layout as novel actuation strategy that redistributes actuator structure circumferentially around an assisted limb. This approach is motivated by applications in lower-limb exoskeletons, where actuators are traditionally positioned in the sagittal plane or mounted remotely at the user’s waist or back. The encircling layout enables the integration of larger diameter actuators without significantly increasing external profile of the device relative to conventional configurations. To facilitate exoskeleton designs with the novel layout, we designed a \textit{Limb-Encircling Actuator} (LEA) and characterized its electromechanical properties on a custom testbed. We also integrated a novel radial bearing assembly into the actuator, termed the \textit{bearing-of-bearings}, that served as lightweight and cost-effective substitute for large-diameter bearings. Overall, this work contributes a design and associated actuation properties that may serve as a reference for exoskeleton designers seeking low-profile, torque-dense actuation strategies. 
    
    % Primary Takeaways
    We established the feasibility of the LEA concept through benchtop experiments, while also identifying key considerations for future design iterations. We designed the LEA using a custom brushless frameless motor that we co-developed for application in an ankle exoskeleton (Section~\ref{design}). We further demonstrated its design viability under different bearing preload configurations via system identification experiments on a custom-built dynamometer. Our design strategy aimed to balance the competing design objectives of utilizing a larger actuator while minimizing its form-factor to facilitate exoskeleton designs that can potentially integrate with clothing and footwear. However, the actuator dimensions were ultimately governed by the internal clearance required for the motor to slide over the foot-ankle complex during donning and doffing. This geometric constrain increased the inner diameter and overall motor dimensions beyond those needed at the final mounted position, preventing the LEA from achieving a close form-fitting envelope. In addition, the adjustability of the bearing-of-bearings helped maintain the required rotor-stator air-gap despite the manufacturing tolerances, but also increased the overall design complexity, mass, and size. Based on these results, future LEA iterations may benefit from eliminating the adjustability features from the bearing-of-bearings design.

    % Secondary Takeaways - Implications of layout on exo design
    The limb-encircling layout may facilitate the development of more torque dense exoskeletons. This layout enables the integration of motors with larger air-gap radii, which directly influence motor torque generation, thermal performance, and steady-state efficiency (Table~\ref{tab:motor_specs}) \cite{seok.kim:2012}. These expected improvements were also observed in the LEA, which exhibited at least a 3.5 fold increase in its torque constant (0.49~Nm/A) relative to the reference motors commonly used in traditional exoskeleton layouts. It also achieved a 3.5 fold increase in its motor constant (0.95~Nm/$\sqrt{\text{W}}$), indicating greater torque output per square root of heat generation. In addition, the LEA dissipated heat more effectively due to its larger surface area, enabling higher continuous and peak currents. As a result, the LEA exhibited a 70.5\% improvement in continuous torque density (7.5~Nm/kg) compared to the reference motors. Consequently, these performance improvements could substantially reduce the transmission ratio required to produce biologically relevant joint torques, potentially lowering the transmission mass and associated losses. 
    
    % Secondary Takeaways Continued - Performance Limitations
    Despite these performance benefits, the limb-encircling geometry introduces trade-offs that affect other aspects of actuator performance. The LEA exhibited a 20 fold increase in rotor inertia relative to the reference motors, which can reduce dynamic-task efficiency and control bandwidth. In wearable robotic systems, electric motors drive dynamic loads and require current to accelerate actuator inertia, resulting in energy loss due to Joule heating of windings. Consequently, higher actuator inertia can significantly increase resistive heat loss during dynamic operation, despite higher efficiency of larger-air gap motors under steady-state conditions. As a result, achieving equivalent runtime may require a larger battery, which in turn could increase overall exoskeleton mass. Moreover, the LEA also achieved a lower position-control bandwidth (7.5~Hz) due to its higher inertia, although this remained sufficient for human locomotion \cite{winter:2009}. In addition, the geometric constraints of donning the actuator prevented the LEA from achieving a close form-fitting envelope (Sec.~\ref{design}). This resulted in a lateral protrusion of 30~mm, which was comparable to the 40~mm protrusion of recent exoskeleton designs \cite{leestma.young:2024, bajpai.mazumdar:2024, zhu.gregg:2021, nesler.gregg:2022}. 

    % Future work
    Future work will integrate the LEA into an ankle exoskeleton and address the existing challenges associated with the limb-encircling layout. The ankle exoskeleton will serve as an initial reference for exploring the implications of transverse plane actuation. In addition, partitioning the actuator structure into smaller modular segments may eliminate the dimensional constraints imposed by donning a rigid hollow cylindrical structure over the distal limb. For example, a design that could open and close the actuator around the limb would avoid the large dimensions required to fit around the foot-ankle complex. Together, these hardware design strategies highlight one strategy for assistive devices that conform to the human body or integrate within clothing or other structured apparel.
      
\section*{Acknowledgment} 
    The authors thank Vamsi Peddinti for his help in developing driver for communication with the brushless motor drives, and Riley Pieper for discussions on brushless motors. They also thank the Consulting for Statistics, Computing, and Analytics Research (CSCAR) group at the University of Michigan for assistance with statistical analysis, and T-Motor (Nanchang, Jiangxi, China) for their support in developing the custom frameless brushless motor used in this work.

    The authors would like to thank T-Motor (Nanchang, Jiangxi, China) for their support in developing the custom frameless brushless motor used in this work. 
    
    AI image generation (OpenAI, San Francisco, CA, USA) was used to create the human silhouette in Fig. \ref{fig:form_factor}.

\bibliographystyle{IEEEtran}
\bibliography{IEEEabrv,references}

\end{document}